%% file: ATC2024_OVQE-VVC.tex
\documentclass[conference]{IEEEtran}
\IEEEoverridecommandlockouts
\usepackage{cite}
\usepackage{amsmath,amssymb,amsfonts}
\usepackage{algorithmic}
\usepackage{graphicx}
\usepackage{textcomp}
\usepackage{xcolor}
\usepackage{multirow}
\usepackage{subcaption}

\def\BibTeX{{\rm B\kern-.05em{\sc i\kern-.025em b}\kern-.08em
    T\kern-.1667em\lower.7ex\hbox{E}\kern-.125emX}}
\begin{document}

\title{Enhancing Quality for VVC Compressed Videos with Omniscient  Quality Enhancement Model \\
}

\author{\IEEEauthorblockN{Xiem HoangVan, Hieu Bui Minh}
\IEEEauthorblockA{\textit{Faculty of Electronics and Telecommunications} \\
{University Of Engineering and Technology}\\
Vietnam National University, Hanoi \\
 {Correspondence: xiemhoang@vnu.edu.vn}
}
\and
\IEEEauthorblockN{Sang NguyenQuang, Wen-Hsiao Peng}
\IEEEauthorblockA{\textit{Department of Computer Science} \\
{National Yang Ming Chiao Tung University}\\
Hsinchu, Taiwan 
}
}

\maketitle

\input{Chapter/0_abstract}
\input{Chapter/1_introduction}
\input{Chapter/2_related_works}
\input{Chapter/3_Proposed_method}
\input{Chapter/4_results}
\input{Chapter/5_conclusions}

\bibliographystyle{IEEEtran}
\bibliography{IEEEfull}

\end{document}

%% file: Chapter/0_abstract.tex
\begin{abstract}
The latest video coding standard H.266/VVC has shown its great improvement in terms of compression performance when compared to its predecessor HEVC standard. Though VVC was implemented with many advanced techniques, it still met the same challenges as its predecessor due to the need for even higher perceptual quality demand at the decoder side as well as the compression performance at the encoder side. The advancement of Artificial Intelligence (AI) technology, notably the deep learning-based video quality enhancement methods, was shown to be a promising approach to improving the perceptual quality experience. In this paper, we propose a novel Omniscient video quality enhancement Network for VVC compressed Videos. The Omniscient Network for compressed video quality enhancement was originally designed for HEVC compressed videos in which not only the spatial-temporal features but also cross-frequencies information were employed to augment the visual quality. Inspired by this work, we propose a modification of the OVQE model and integrate it into the lasted STD-VVC (Standard Versatile Video Coding) decoder architecture. As assessed in a rich set of test conditions, the proposed OVQE-VVC solution is able to achieve significant PSNR improvement, notably around 0.74 dB and up to 1.2 dB with respect to the original STD-VVC codec. This also corresponds to around 19.6\% of bitrate saving while keeping a similar quality observation. 
\end{abstract}

%% file: Chapter/1_introduction.tex
\section{Introduction}
\label{sec:intro}
Nowadays, video content has been expanded into various formats, including ultra-high resolutions (4K, 8K), 360-degree videos, as well as augmented reality (AR) and virtual reality (VR) experiences. In this context, the need for a video codec that provides better quality and has a wide range of applications is the main motivation behind the development of Versatile Video Coding (VVC)\cite{overview_VVC}. The VVC standard aims to reduce video compression bitrates by up to half compared to its predecessor, HEVC \cite{overview_HEVC} and serves as a unified coding method for multiple types of video. Additionally, VVC is a versatile encoder with integrated tools that support multiview and scalable encoding \cite{SVVC}. For both video standards, the primary research focuses on two main directions: enhancing encoding performance and reducing computational complexity \cite{FastSearch,Trellis}. In addition, with the advancement of artificial intelligence across various fields \cite{PhobertDLModel}, deep learning has been employed to improve the performance of video encoders.

\begin{figure}[tb]
    \centering
    \centerline{\includegraphics[width=0.5\textwidth]{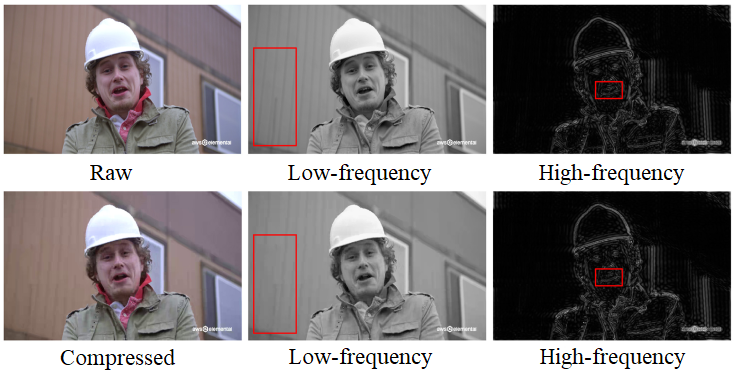}}
    \caption{Loss of high-frequency information due to compression \cite{peng2022ovqe}.}
    \label{fig:high_freq_loss}
\end{figure}

VVC still follows a block-based video-like codec in which frames are divided into blocks called coding units (CUs), and calculations are performed in the cosine-transformed domain. One of the remaining issues with this video coding approach is the loss of the high-frequency information, which occurs due to performing quantization in the frequency domain \cite{AdaptiveQP}, as shown in  Fig. \ref{fig:high_freq_loss}. It can be seen that, due to the compression impact, there is a reduction in the quality of frames. Subsequently, this causes inconsistency in the experience of media content consumers, resulting in a reduction of Quality of Experience (QoE).

To address this issue, quality enhancement (QE) methods were introduced. Typically, the QE models are designed to exploit the high correlation between consecutive decoded frames at the decoder sides. In this approach, either the "loss" high-frequency information is restored using a probabilistic model \cite{rao2012survey} or using a deep learning approach \cite{CF-STIF,STDR,TVQE,peng2022ovqe}. Among the deep learning-based approach, omniscient Network for Compressed Video Quality Enhancement (OVQE)\cite{peng2022ovqe}, which contains the Omni-Frequency Adaptive Enhancement (OFAE) Block, is introduced to learn the omni-frequency information of compressed video adaptively, thereby improving the quality of the reconstructed video, which is encoded by HEVC standard. However, the original OVQE has not been examined for the newly VVC standard, its model needs to have a comprehensive analysis and verification. 

In this work, to achieve high perceptual quality for VVC decoded pictures, we propose a novel OVQE\_VVC architecture in which the OVQE model is utilized to enhance the quality of reconstructed video encoded by VVC standard. The OVQE impaction together with its limitation are carefully analyzed in the landscape of VVC decoded information. An extensive performance evaluation is carried out to compare the proposed OVQE\_VVC with the standard VVC and the most relevant MFQE 2.0\cite{guan2019mfqe} and STDF quality enhancement model\cite{stdf}. Experimental results show that the proposed OVQE\_VVC solution can bring around 0.74dB to 1.2dB of quality improvement for decoded video from several landscapes. Indirectly, it shows that the proposed OVQE\_VVC codec outperforms the standard VVC codec with around 19.6 \% bitrate saving while providing a similar perceptual quality.

To demonstrate this work, the rest of this paper is organized as follows. Section II briefly presents the related works on video quality enhancement. Meanwhile, Section III describes the proposed OVQE\_VVC framework as well as the construction of the OVQE model. Afterward, Section IV presents the quality and compression efficiency assessments for the proposed video codec. Finally, Section V gives some conclusions and outlines future works. 
 

%% file: Chapter/2_related_works.tex
\section{Related works}
\label{sec:related_works}


Recently, deep learning-based methods proved to be the way to go for the future of video quality enhancement. The two main approaches to deep learning quality enhancement are the single-frame-based method and the multi-frame-based method. The first one attempts to exploit the spatial redundancy as presented in \cite{yang2017decoder, yang2018enhancing, zhang2017beyond, wang2017novel}. Multi-frame-based methods, which show superior results, utilize the spatial-temporal information of compressed video to reconstruct its original quality. Many works have been made using multi-frame-based methods to improve the quality of decoded HEVC video. Yang~\emph{et al.} proposed a Multi-Frame Quality Enhancement (MFQE) \cite{yang2018multi} model with a Support Vector Machine (SVM) classifier to detect Peak Quality Frames (PQFs) and use a Motion-Compensation subnet (MC-subnet) to compensate the temporal motion between neighboring frames. Later, Guan~\emph{et al.} create a improved model called MFQE 2.0 \cite{guan2019mfqe} by using a Multi-Frame Convolutional Neural Network (MF-CNN) and a Bidirectional Long Short-Term Memory (BiLSTM) detector. MFQE 2.0 also showed quality improvement with VVC decoded sequence \cite{hoangvan2020enhancing}. Luo~\emph{et al.}\cite{CF-STIF}  proposed to use 3D convolutional, which has a large spatiotemporal receptive field, to obtain spatiotemporal features and design the multi-level residual fusion module to fuse global and local features. Wang~\emph{et al.}\cite{STIB} proposed a plug-and-play module to balance spatial and temporal information. Li~\emph{et al.}\cite{TVQE} applied transformer at multi-scale manner to learn local and global features.  Luo~\emph{et al.}\cite{STDR} proposed a recurrent deformable fusion method that aligns each pair of target and adjacent frames sequentially along the timeline, using the compensation information between adjacent frames more efficiently than aligning multiple frames simultaneously.

\begin{table*}[htb]
\caption{Summary of recent works on video quality enhancement}
\begin{tabular}{|p{15mm}|p{50mm}|p{50mm}|p{50mm}|}
\hline
\centering
\textbf{Method}     & \textbf{Key ideas} & \textbf{Advantages}         & \textbf{Limitations}                                                      \\ \hline
MFQE2.0 (2019)   & Detect Peak-Quality-Frame (PQF) frames, and then use information from PQF frames to enhance the quality of non-PQF frames     & Leverages the quality fluctuation across frames, effectively enhancing low-quality frames with information from neighboring high-quality frames                                                                     & The networks' excessively large parameters challenge training efficiency and real-time tasks, while existing methods prioritize enhanced results at the expense of inference speed.                                                             \\ \hline
STDF (2020)      & Uses a novel approach that involves multi-frame deformable convolutions to align features without explicit motion estimation.     & Significantly enhances compressed video quality by effectively aggregating temporal information through multi-frame deformable convolutions, leading to superior accuracy and efficiency                                                                    & Not utilize past and future enhanced results                    \\ \hline
RFDA (2021)       & Employs a recursive fusion approach to merge spatiotemporal features from various frames and uses deformable spatiotemporal attention to prioritize areas with high artifacts     & Utilizing spatiotemporal information through recursive fusion and deformable attention                                                                    & Not exploit the spatiotemporal information of the entire video  \\ \hline
BasicVSR++ (2022 )& Incorporating second-order grid propagation and flow-guided deformable alignment     & Utilize spatiotemporal information effectively, leading to significantly improved performance over existing methods while maintaining similar computational demands                                                                    & Current frame cannot utilize all information of adjacent frames \\ \hline
OVQE (2022)       & Perform bi-directional propagation to leverage all of the information from adjacent frames. Afterward, perform quality enhancement in frequency domain     & Fully utilized the omni spstiotemporal correlations of the whole video & Performance depends on the number of bidirectional propagation round.                                                             \\ \hline
\end{tabular}
\end{table*}

One of the most advanced models for quality enhancement recently is the Omniscient Network for Compressed Video Quality Enhancement (OVQE) \cite{peng2022ovqe}. OVQE is an improvement of BasicVSR++\cite{chan2022basicvsr++} and OVSR\cite{yi2021omniscient} and built for MFQE 2.0 dataset encoded with HEVC Low-Delay-P configuration. OVQE adopts many novel techniques such as deformable convolutional neural network (deformable CNN)\cite{dai2017deformable} and omni-frequency domain enhancement. With the promising and advanced architecture of OVQE, we proposed a video coding framework with VVC encoder-decoder and an OVQE postprocessing block for quality enhancement. The experimental result archived proves the effectiveness of our framework.

%% file: Chapter/3_proposed_method.tex
\section{Proposed Method}
In this section, we present a learning-based approach to enhance the quality of VVC reconstructed video, namely OVQE\_VVC. OVQE, previously applied to HEVC, increased PSNR by around 1.2 dB and achieved state-of-the-art quality enhancement. Despite VVC's improved visual quality and reduced bitrate, compression artifacts remain. Therefore, a quality enhancement method for VVC is necessary. We leverage OVQE's proven capability to address this issue, as no current solution exists for the latest VVC standard.
\label{sec:proposed_method}
\subsection{Overall OVQE\_VVC Framework}
\label{ssec:ovqe_vvc}
\begin{figure*}[t!]
    \centerline{\includegraphics[width=0.75\textwidth]{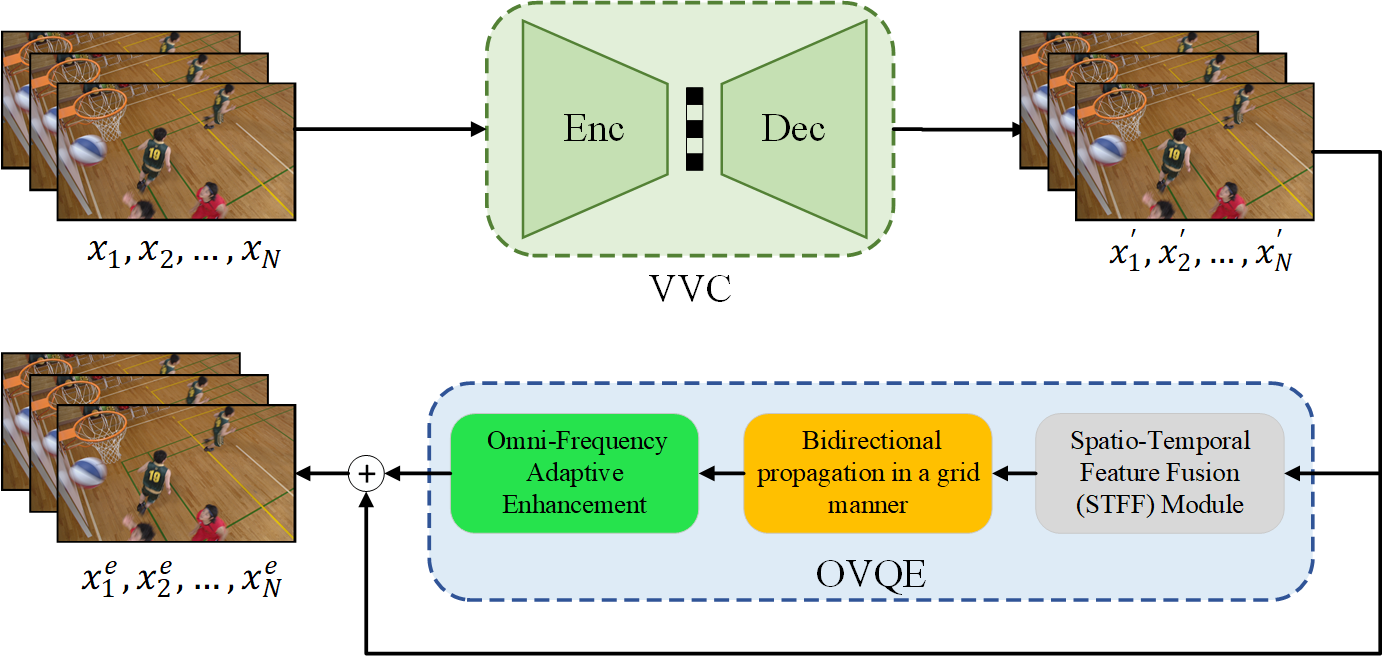}}
    \caption{Illustration of our proposed method. }
    \label{fig:propose_ovqe_vvc}
    \vspace{-1.5em}
\end{figure*}
Our proposed approach is illustrated in Fig \ref{fig:propose_ovqe_vvc}.  We focus on improving the compressed video coded by recent VVC standard. Given a video sequence containing $N$ frames $\left \{ x_{1},x_{2},...,x_{N} \right \}$ where $x_{i}$ is the frame at time $i$. At the encoder side, the raw video goes through VVC to be the compressed video with size reduction. The compressed bitstream is then decoded at the decoder side to obtain reconstructed video $\left \{ x_{1}^{'},x_{2}^{'},...,x_{N}^{'} \right \}$. This process can be represented as:
\begin{equation}
\label{equ:enc_dec_vvc}
\left \{x_{1}^{'},x_{2}^{'},...,x_{n}^{'}\right \}=f_{dec}^{VVC}\left ( f_{enc}^{VVC}\left ( x_{1},x_{2},...,x_{n} \right ) \right ),
\end{equation} 
where $f_{enc}^{VVC}(\cdot)$ and $f_{dec}^{VVC}(\cdot)$ represent the encoder and decoder of VVC standard, respectively. 

Note that the compression artifacts adversely affect the quality of reconstructed frames, especially the loss of high-frequency details. To obtain higher visual quality video at the decoder side, we adopt a video quality enhancement method to reduce compression artifacts, which can be expressed as:
\begin{equation}
\label{equ:ovqe}
\left \{x_{1}^{e},x_{2}^{e},...,x_{n}^{e}\right \}=OVQE\left ( x_{1}^{'},x_{2}^{'},...,x_{n}^{'} \right ),
\end{equation}
where $OVQE(\cdot)$ represents the Omniscient Network for Compressed Video Quality Enhancement \cite{peng2022ovqe}, $\left \{x_{1}^{e},x_{2}^{e},...,x_{n}^{e}\right \}$ represent high-quality video frame.

\subsection{OVQE Model}
\label{ssec:ovqe}
Omniscient Network is designed to learn the video spatiotemporal and omni-frequency information in an effective way. As presented in Fig. \ref{fig:propose_ovqe_vvc}, OVQE contains 3 main blocks: Spatio-Temporal Feature Fusion (STFF) Module,  Bidirectional propagation in a grid manner and Omni-Frequency Adaptive Enhancement block.

Given a frame $x_{t}, $ has a low quality that needs to be enhanced, the OVQE model utilizes the spatiotemporal information of $R$ adjacent frames in the past and $R$ adjacent frames in the future. Firstly, the Spatio-Temporal Feature Fusion (STFF) module, which consists of SKU-Net and deformable convolution, is used to extract information at different scales and perform fusion. This step can be written as follows:
\begin{equation}
\label{equ:stff}
f_{t}=STFF\left ( \left \{ x_{1}^{'},x_{2}^{'},...,x_{n}^{'} \right \} , {\phi}_{s}  \right),
\end{equation}
where $\phi_{s}$ are the the learnable parameters and $f_{t}$ is the fused spatial-temporal feature.

\begin{figure}[htb]
    \centering
    \centerline{\includegraphics[width=0.45\textwidth]{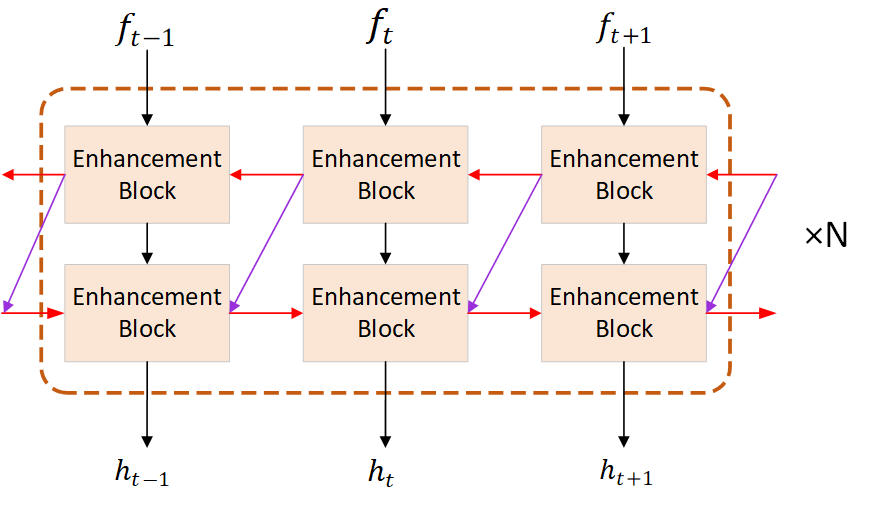}}
    \caption{Bidirectional propagation in a grid manner}
    \label{fig:bi-propagation}
\end{figure}

Afterward, the fused feature is used to perform bidirectional propagation in a grid manner. Fig. \ref{fig:bi-propagation} illustrates how the bidirectional propagation in a grid manner mechanism works. In this stage, backward propagation is performed first, followed by forward propagation. Assume that the extracted feature of the current frame is $f_{t}$. During backward propagation, the model uses the hidden state $f_{t+1}$ as the input to align the features at the current hidden state. Then, the enhanced features of past, present, and future are input to get the aligned feature. Both backward and forward propagation processes use the Enhancement Block, containing STFF and OFAE block, to generate the hidden state $h_{t}$. This process can be repeated $N$ rounds, but the greater the number of rounds, the greater the complexity. 

After performing bidirectional propagation in a grid manner, the enhanced future $h_{t}$ is obtained and used to perform omni-frequency enhancement. In this stage, several $OFAE$ blocks are used to enhance the future in frequency domain. Omni-Frequency Feature Extraction is performed first to extract low, mid, and high-frequency components. Then, these frequency components are enhanced and fused by Omni-Frequency Feature Enhancement block.

\subsection{Training Strategy}
\label{ssec:training_strategy}
OVQE framework is trained by using MFQE2.0 dataset. During the training phase, the Charbonnier Loss \cite{Charbonnier_Loss}, as expressed in Eq. \ref{equ:Charbonnier_loss} and Adam optimizer \cite{Adam} are used to optimize the model.
\begin{equation}
\label{equ:Charbonnier_loss}
loss=\frac{1}{n}\sum_{i=1}^{n}\sqrt{\left ( x_{i}-\hat{x}_{i} \right )^2+\epsilon },
\end{equation}
where $n$ is the number of frames of input video,
$x_{i}$ and $\hat{x}_{i}$ denote the ground truth and the frame generated by the OVQE network, respectively. $\epsilon$ is set as $10^{-6}$ 

%% file: Chapter/4_results.tex
\section{Experiment Results	}
\label{sec:results}
\subsection{Test Condition}
\label{ssec:test_condition}
\begin{figure}[t!]
\centering
\footnotesize
{
\begin{tabular}{ccc}

    \includegraphics[width=0.3\linewidth]{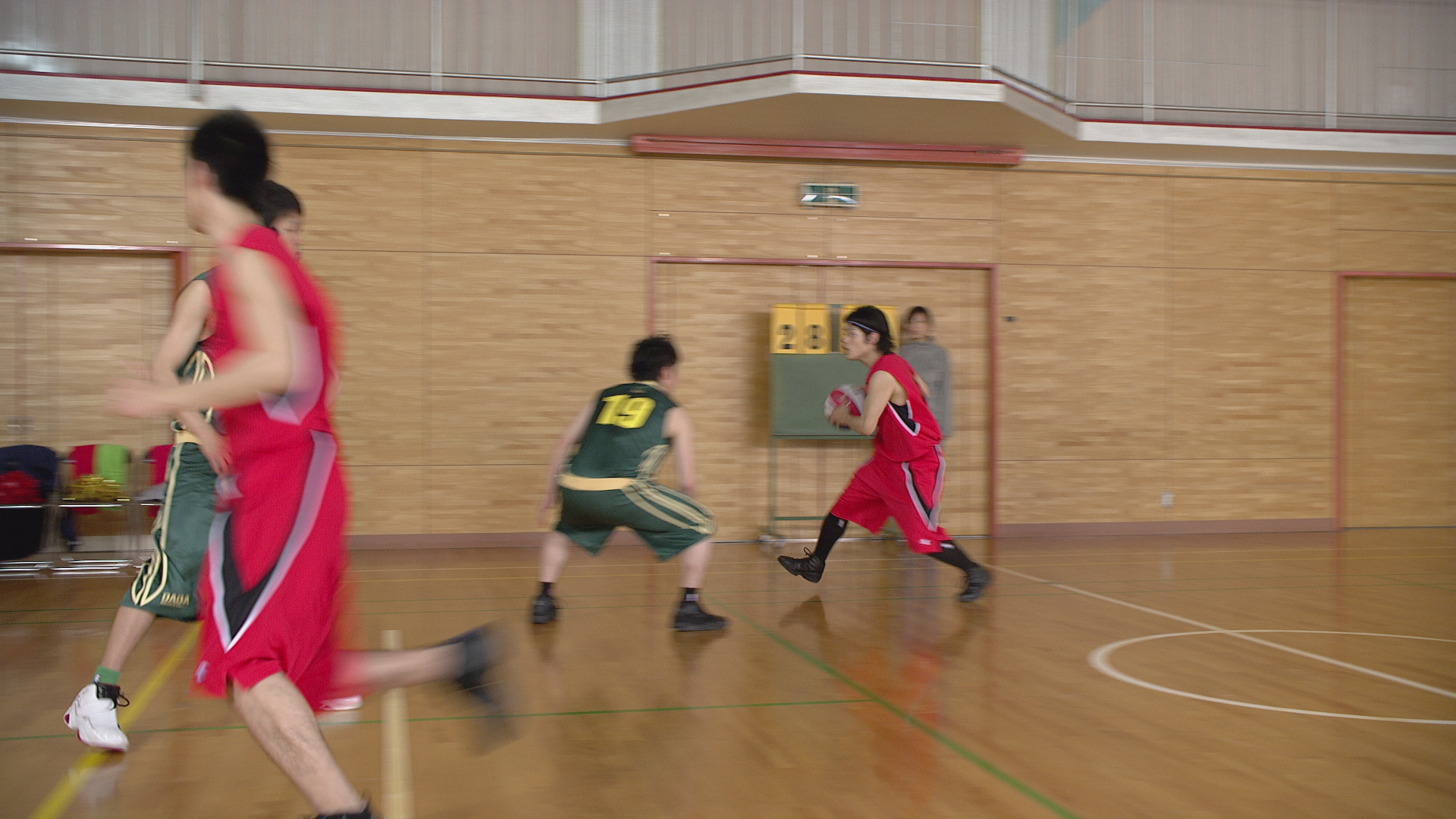} &
    \includegraphics[width=0.3\linewidth]{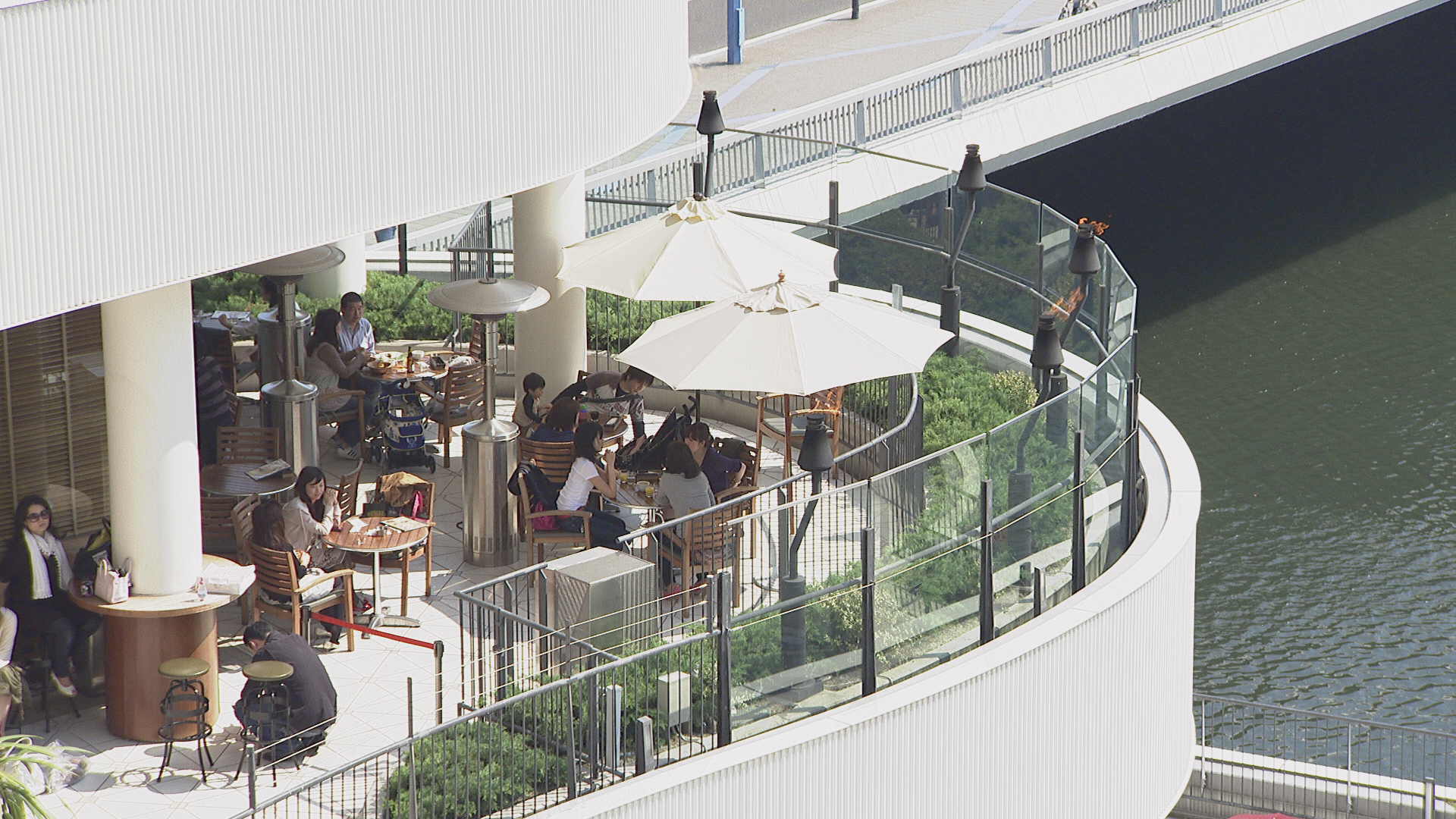} &
    \includegraphics[width=0.3\linewidth]{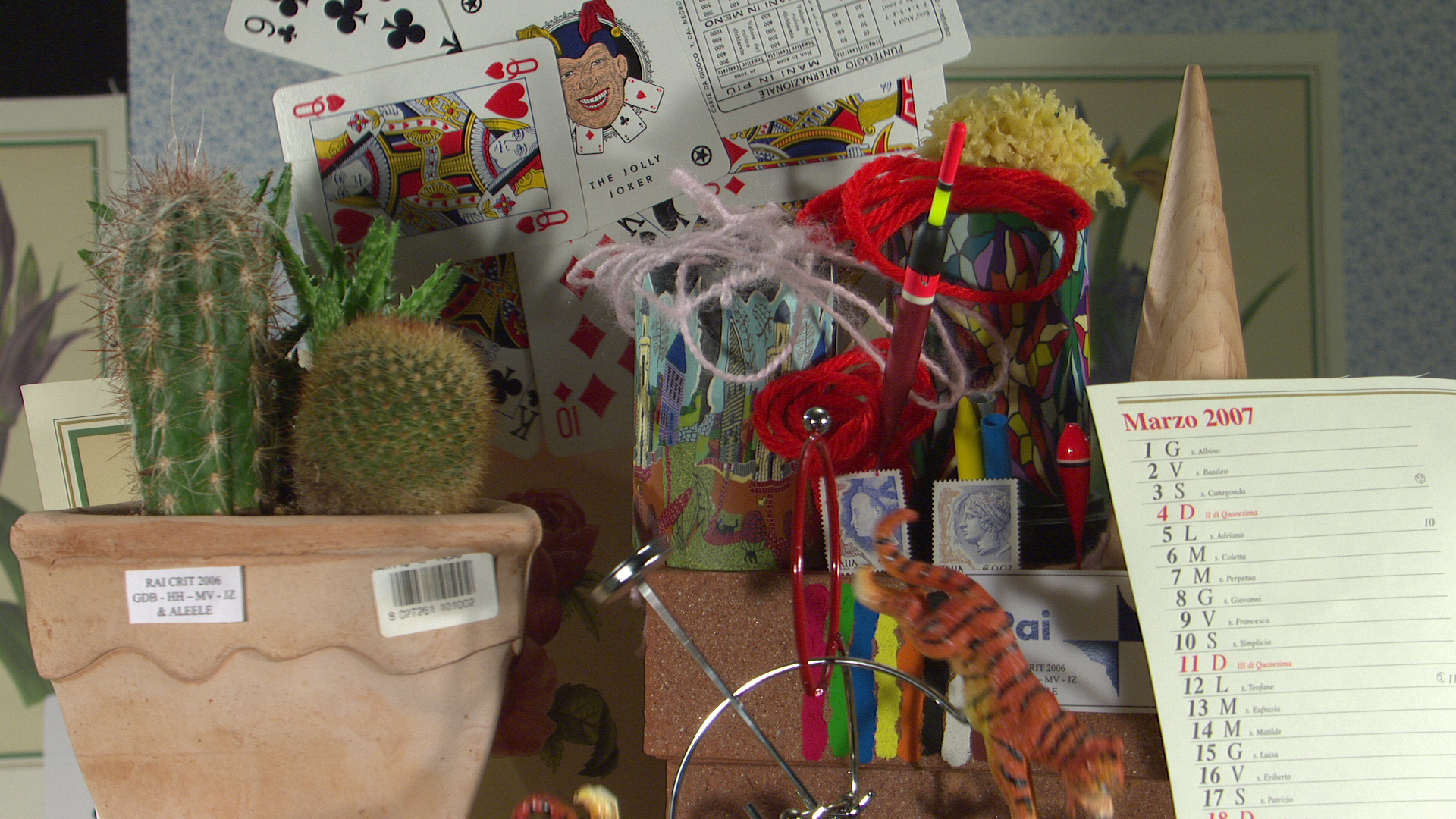} \\
     \multicolumn{1}{c}{BasketballDrive} &  \multicolumn{1}{c}{BQTerrace} &  \multicolumn{1}{c}{Cactus}  \\
     
    \includegraphics[width=0.3\linewidth]{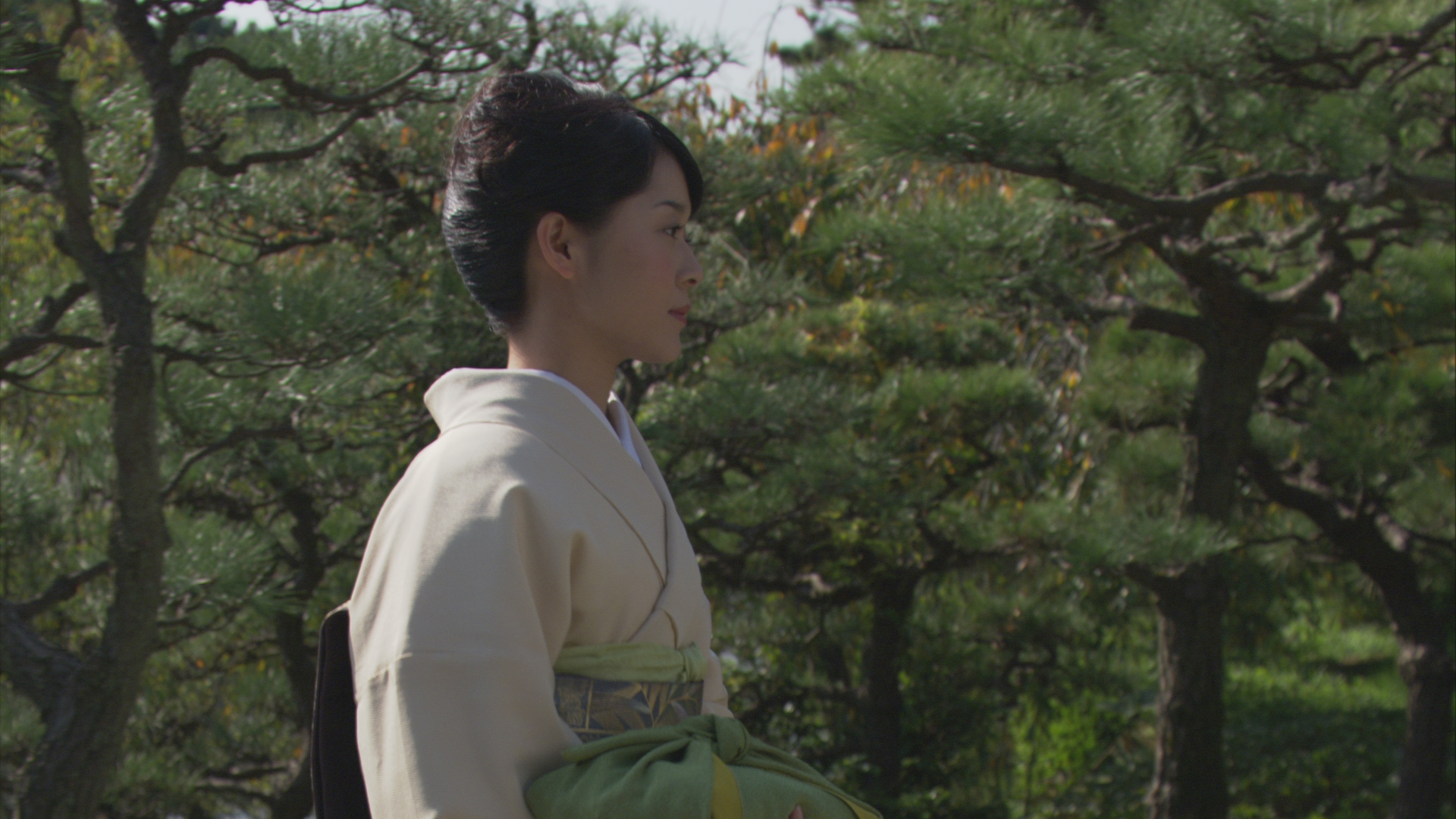} &
    \includegraphics[width=0.3\linewidth]{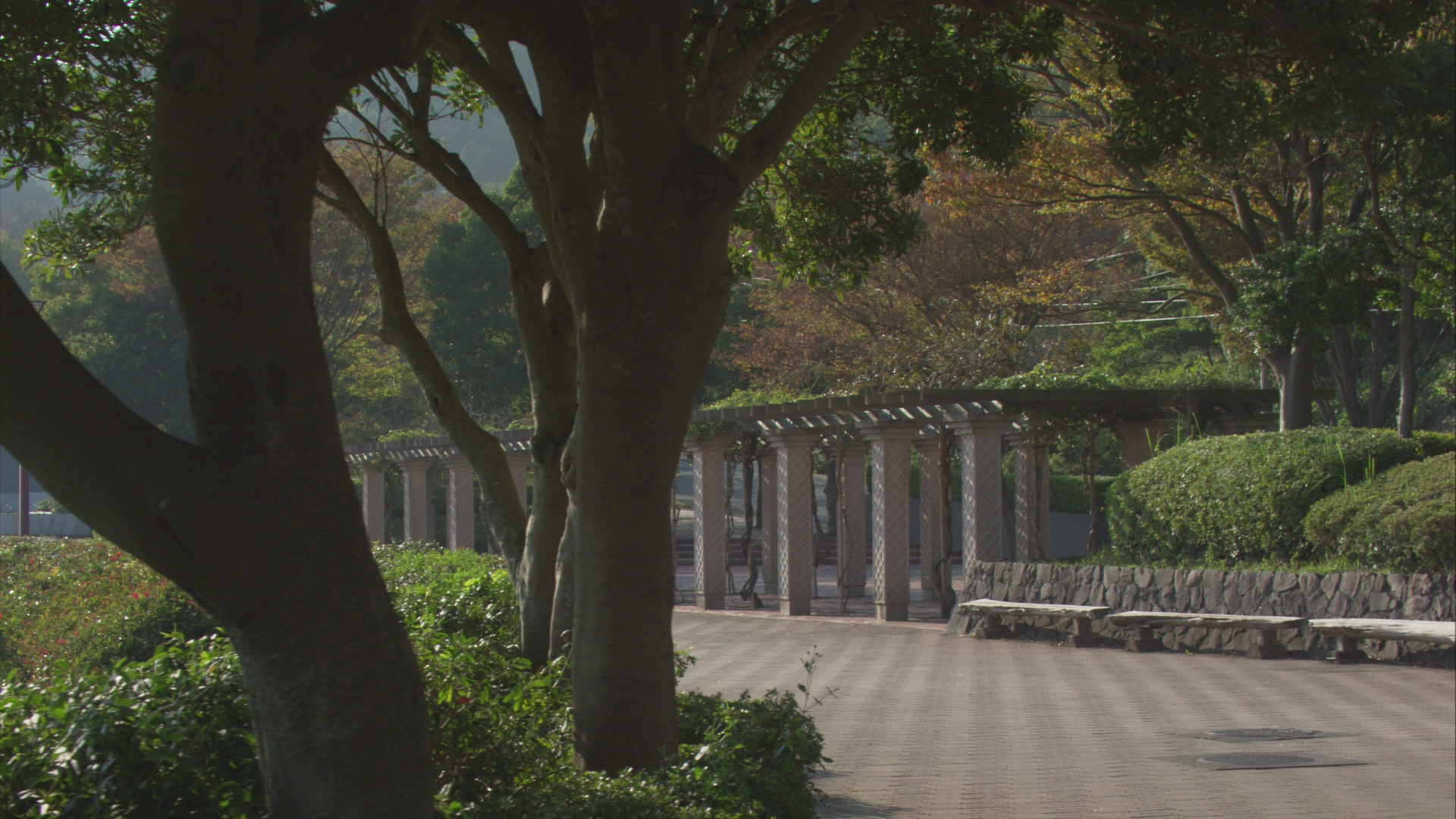} &
    \includegraphics[width=0.3\linewidth]{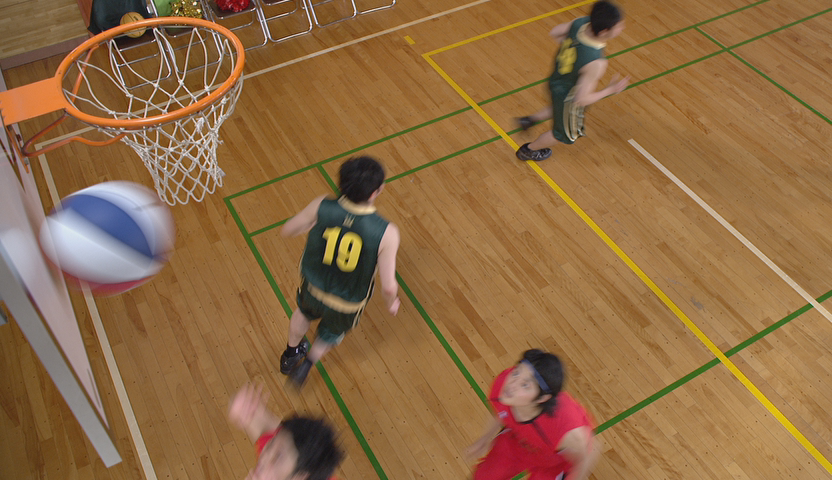} \\
     \multicolumn{1}{c}{Kimono} &  \multicolumn{1}{c}{ParkScene} &  \multicolumn{1}{c}{BasketballDrill}  \\

    \includegraphics[width=0.3\linewidth]{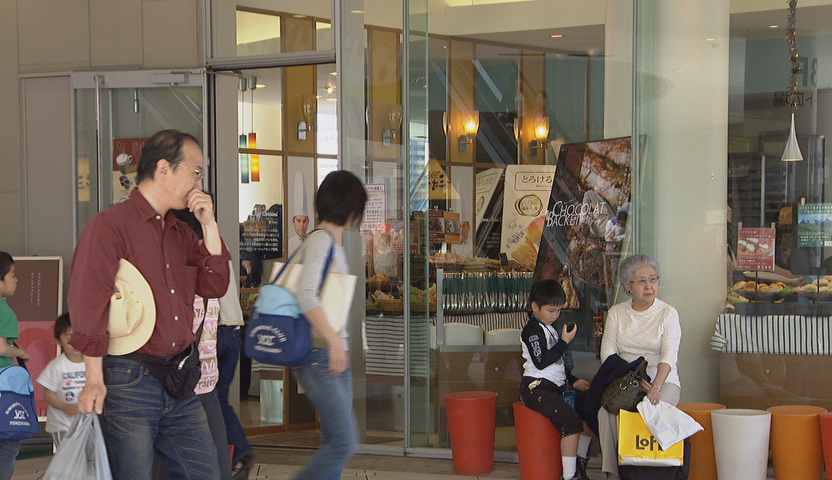} &
    \includegraphics[width=0.3\linewidth]{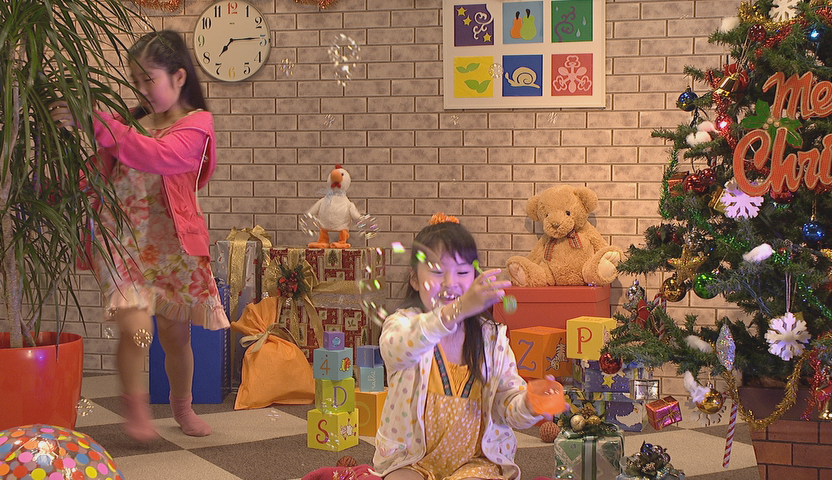}     &
    \includegraphics[width=0.3\linewidth]{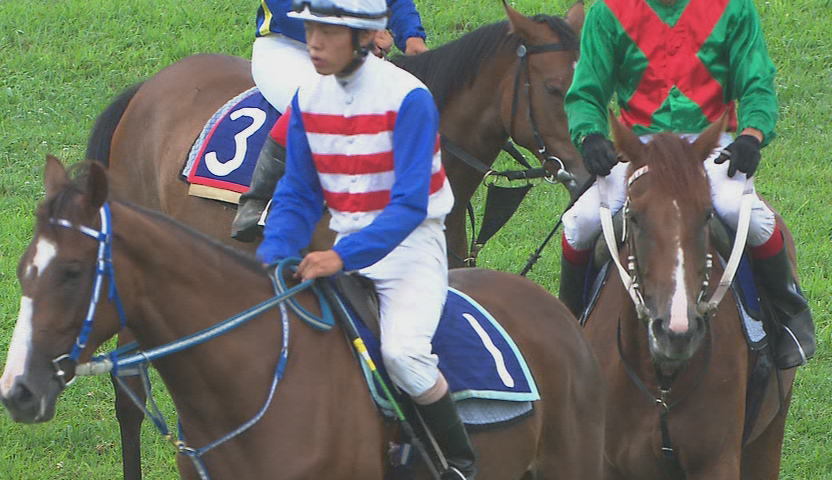}  \\
     \multicolumn{1}{c}{BQMall} &  \multicolumn{1}{c}{PartyScene} &  \multicolumn{1}{c}{RaceHorses}  \\

    \includegraphics[width=0.3\linewidth]{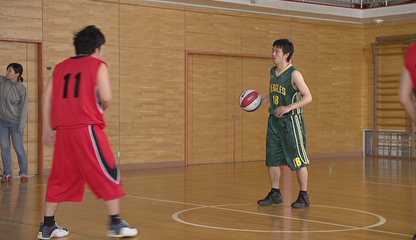} &
    \includegraphics[width=0.3\linewidth]{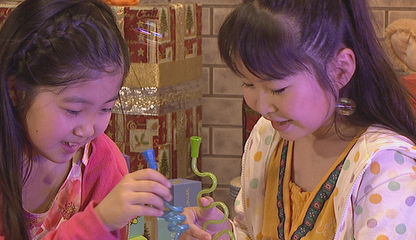}      &
    \includegraphics[width=0.3\linewidth]{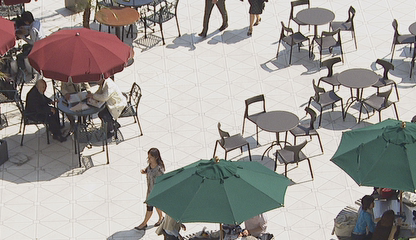} \\

     \multicolumn{1}{c}{BasketballPass} &  \multicolumn{1}{c}{BlowingBubbles} &  \multicolumn{1}{c}{BQSquare}  \\

\end{tabular}
}

\vspace{-1mm}

\caption{First frame of test video sequences.}

\label{fig:first_frame}
\end{figure}

In order to evaluate the performance of the proposed OVQE\_VVC approach, we use VVenC \cite{VVenC} to compress 13 video sequences from VVC Common Test Conditions (CTC) \cite{vvc_ctc} with different setting the Quantization Parameters (QPs) to 32, 37, 42 and 47, and then use OVQE to enhance the quality of reconstructed video at the decoder side. The first frame of test video sequences is illustrated in Fig. \ref{fig:first_frame}.

We measure $\Delta PSNR$, as expressed in Eq. \ref{equ:delta_psnr} to show the quality enhancement performance of using the proposed approach.
\begin{equation}
\label{equ:delta_psnr}
\Delta PSNR=PSNR_{OVQE\_VVC}-PSNR_{STD\_VVC},
\end{equation}
where $PSNR_{STD\_VVC}$ is the quality of the decoded video using VVC standard, while $PSNR_{OVQE\_VVC}$ represents the quality of the decoded video after performing enhancement using OVQE.

\subsection{Quality Enhancement Assessment}
\label{ssec:performance_evaluation}
\begin{figure}[htb]
    \centering
    \centerline{\includegraphics[width=0.5\textwidth]{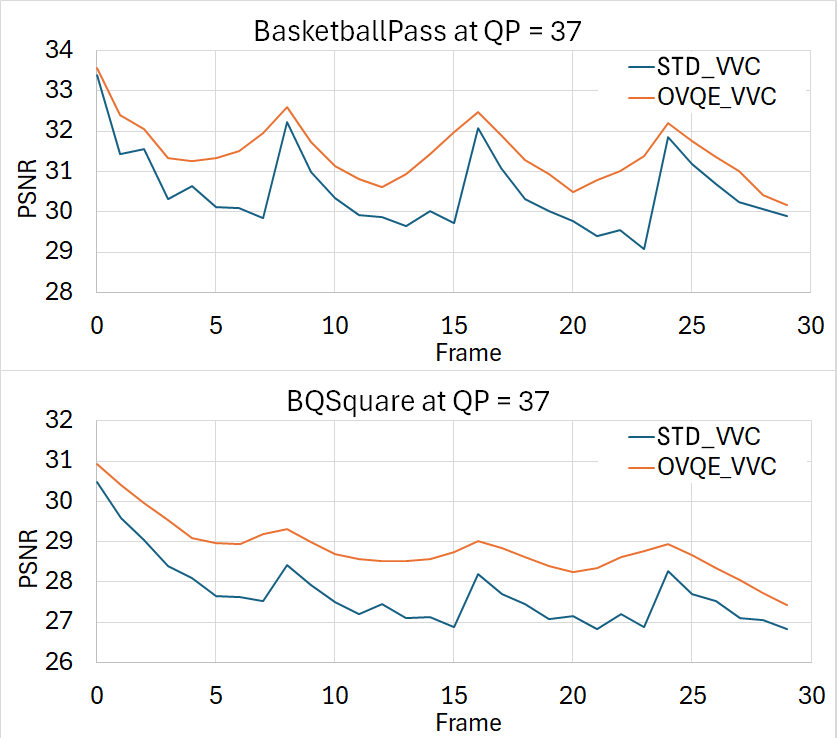}}
    \caption{PSNR curves of VVC baseline and proposed OVQE\_VVC approach.}
    \label{fig:psnr_per_frame}
\end{figure}

Table \ref{tab:psnr_results} shows the average quality improvement measured  with the detal PSNR [dB] while Fig. \ref{fig:psnr_per_frame} shows the quality comparison frame by frame with the proposed OVQE\_VVC with respect to the standard VVC solution. As shown, the proposed OVQE solution significantly improves the quality of reconstructed pictures. 

To further assess the quality enhancement with OVQE, we measured and compared with also two most relevant QE methods, notably the MFQE version 2 \cite{guan2019mfqe} and the STDF \cite{stdf}. As shown in Table II. Except for the BasketballDrill at QP 32, the OVQE model outperforms both MFQE and STDF approaches. This mainly comes from the use of both spatio-temporal and frequency information as detailed in the proposed Section III. 

\begin{table}[htb]
\caption{PSNR gain with proposed approach.}
\begin{tabular}{cccccc}
\hline
\multirow{2}{*}{\textbf{Class}}   & \multirow{2}{*}{\textbf{Sequence}} & \multicolumn{4}{c}{\textbf{Delta PSNR}}                            \\ \cline{3-6} 
                         &                           & \textbf{QP 47}        & \textbf{QP 42}        & \textbf{QP 37}        & \textbf{QP 32}        \\ \hline
\multirow{5}{*}{Class B} & BasketballDrive           & 0.283        & 0.480        & 0.536        & 0.268        \\
                         & BQTerrace                 & 0.175          & 0.274          & 0.271          & 0.081          \\
                         & Cactus                    & 0.218          & 0.394          & 0.457          & 0.252          \\
                         & Kimono                    & 0.239        & 0.434          & 0.569        & 0.331        \\
                         & ParkScene                 & 0.135        & 0.256        & 0.404        & 0.401        \\ \hline
\multirow{4}{*}{Class C} & BasketballDrill           & 0.334        & 0.542        & 0.153        & -0.422       \\
                         & BQMall                    & 0.321        & 0.680        & 0.996        & 0.710        \\
                         & PartyScene                & 0.165        & 0.401        & 0.658        & 0.600        \\
                         & RaceHorsesC               & 0.219        & 0.303        & 0.351        & 0.286        \\ \hline
\multirow{4}{*}{Class D} & BasketballPass            & 0.298        & 0.672        & 1.099        & 0.923        \\
                         & BlowingBubbles            & 0.283        & 0.561        & 0.950        & 0.982        \\
                         & BQSquare                  & 0.343        & 0.822        & 1.232        & 1.148        \\
                         & RaceHorses                & 0.257        & 0.504        & 0.671        & 0.568        \\ \hline
\multicolumn{2}{c}{\textit{Average}}                 & \textit{0.252} & \textit{0.486} & \textit{0.642} & \textit{0.471} \\ \hline
\label{tab:psnr_results}
\end{tabular}
\end{table}

\begin{table}[htb]
\caption{Delta PSNR improvement with various QE models}
\centering
\begin{tabular}{lcccc}
\hline
\textbf{Sequence}                         & \textbf{QP} & \textbf{MFQE\_VVC}   & \textbf{STDF\_VVC} & \textbf{OVQE\_VVC}   \\ \hline
\multirow{2}{*}{BasketballDrill} & 37  & 0.142  & 0.385     & 0.153  \\
                                 & 32  & 0.176  & -0.073    & -0.422  \\ \hline
\multirow{2}{*}{BQMall}          & 37  & 0.592  & 0.694     & 1.232  \\
                                 & 32  & 0.260  & 0.558     & 1.148  \\ \hline
\multirow{2}{*}{PartyScene}      & 37  & 0.196  & 0.424     & 0.658  \\
                                 & 32  & 0.264  & 0.378     & 0.600  \\ \hline
\multirow{2}{*}{RaceHorsesC}     & 37  & 0.068  & 0.266     & 0.351  \\
                                 & 32  & 0.046  & 0.075     & 0.286  \\ \hline
\multirow{2}{*}{BQSquare}        & 37  & 0.223 & 0.551     & 0.710  \\
                                 & 32  & 0.242  & 0.486     & 1.000  \\ \hline
\multirow{2}{*}{BasketballPass}  & 37  & 0.297  & 0.704     & 1.099  \\
                                 & 32  & 0.299  & 0.583     & 0.923  \\ \hline
\multirow{2}{*}{BlowingBubbles}  & 37  & 0.410   & 0.498     & 0.950  \\
                                 & 32  & 0.317  & 0.515     & 0.982  \\ \hline
\multirow{2}{*}{RaceHorses}      & 37  & 0.169  & 0.496     & 0.671  \\
                                 & 32  & 0.169  & 0.387     & 0.568  \\ \hline
\multicolumn{2}{c}{\textit{Average}}    & \textit{0.242}   & \textit{0.422} & \textit{0.682}   \\ \hline
\end{tabular}
\label{tab:delta_PSNR}
\end{table}

\begin{figure}[t]
    \begin{center}
    \begin{subfigure}{0.45\linewidth}
        \centering
        \includegraphics[width=\linewidth]{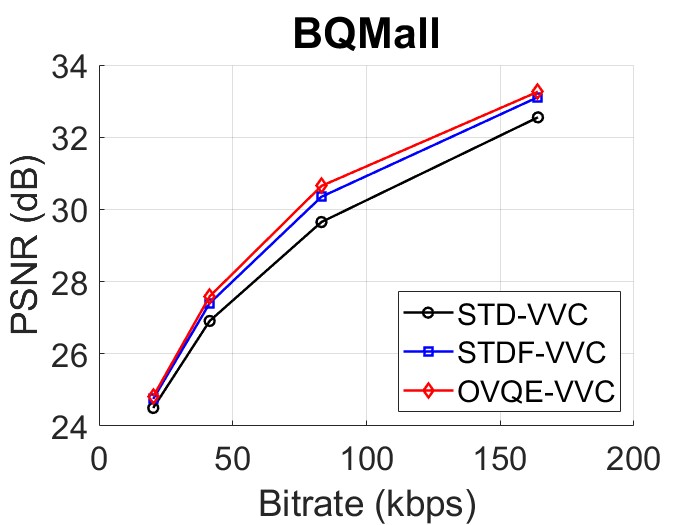}
        \label{fig:RD_BQMall}
    \end{subfigure}
    \begin{subfigure}{0.45\linewidth}
        \centering
        \includegraphics[width=\linewidth]{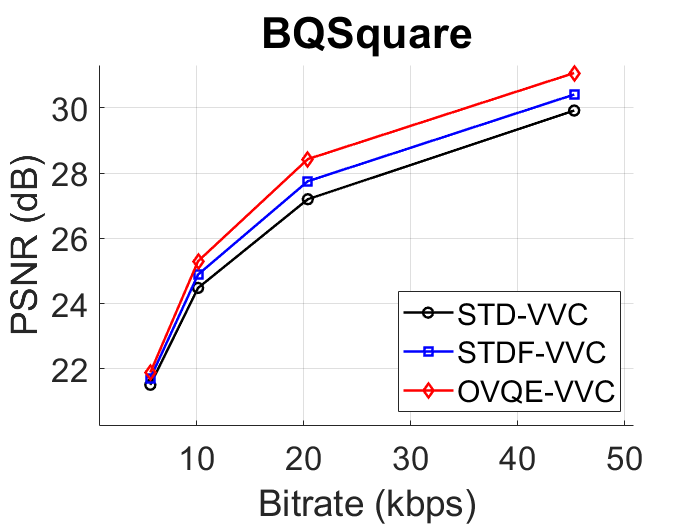}
        \label{fig:RD_BQSquare}
    \end{subfigure}
    \begin{subfigure}{0.45\linewidth}
        \centering
        \includegraphics[width=\linewidth]{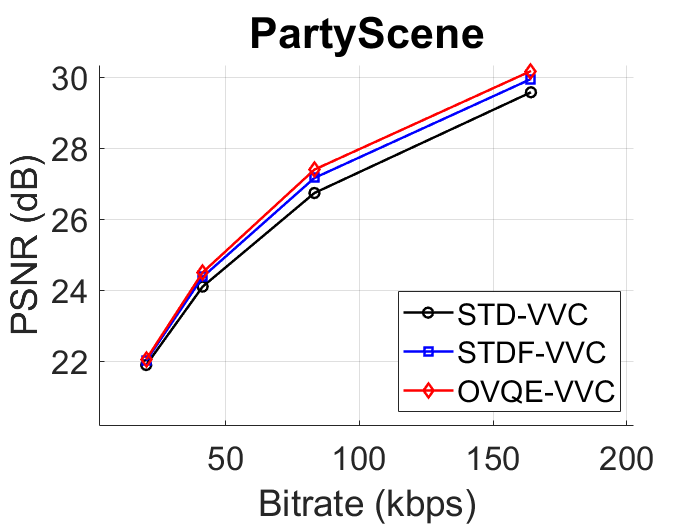}
        \label{fig:RD_partyscene}
    \end{subfigure}
        \begin{subfigure}{0.45\linewidth}
        \centering
        \includegraphics[width=\linewidth]{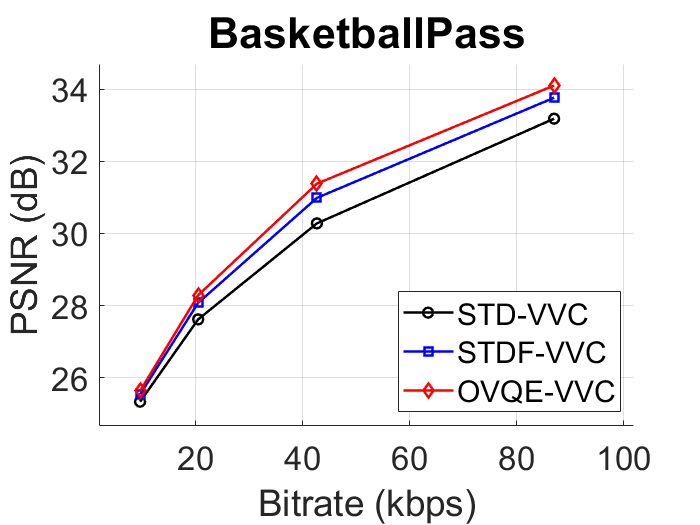}
        \label{fig:RD_basketballpass}
    \end{subfigure}

    \caption{RD comparison.}
    \label{fig:RD_curve}
    \end{center}
    \vspace{-1.5em}
\end{figure}

\begin{table}[tb]
\centering
\caption{BD-Rate of OVQE\_VVC and STDF\_VVC compare to STD\_VVC}
\begin{tabular}{cccc}
\hline
\textbf{Class}   & \textbf{Sequence} & \textbf{STDF\_VVC}     & \textbf{OVQE\_VVC}  \\ \hline
\multirow{4}{*}{Class C} & BasketballDrill           & -10.559        & -10.260        \\
                         & BQMall                    & -14.845        & -21.090        \\
                         & PartyScene                & -12.434        & -19.170        \\
                         & RaceHorsesC               & -7.758         & -14.680        \\ \hline
\multirow{4}{*}{Class D} & BasketballPass            & -15.692        & -24.420        \\
                         & BlowingBubbles            & -15.172        & -29.690        \\
                         & BQSquare                  & -11.757        & -22.690        \\
                         & RaceHorses                & -14.522        & -15.110        \\ \hline
\multicolumn{2}{c}{\textit{Average}}                 & -12.538        & -19.639         \\ \hline
\label{tab:BD-Rate}
\end{tabular}
\end{table}

\subsection{RD Performance Assessment}
\label{ssec:RD_performance_evaluation}

Finally, to assess the compression performance in general when ultilizing the QE methods. We measured and compared the RD performance of the proposed OVQE\_VVC with the standard VVC and STDF\_VVC. While the RD performance curves are illustrated in Fig. 6, the BD-Rate improvements are shown in Table \ref{tab:delta_PSNR}. 

As it can be seen, the proposed OVQE\_VVC significantly outperforms both standard STD\_VVC and the relevant STDF\_VVC. Especially, the BD-Rate improvement can be up to -29.690 while keeping a similar decoded perceptual quality. This again confirms that the proposed OVQE model not only directly improves the perceptual quality at the decoder but also indirectly improves the compression performance for VVC compression approach. 

%% file: Chapter/5_conclusions.tex
\section{Conclusions}
\label{sec:conclusions}

In this paper, to improve the quality of decoded video obtained from the VVC compression solution, we propose and integrate a new Omniscient Video Quality Enhancement model for the VVC decoder. Difference to prior quality enhancement models, the OVQE model exploits not only the spatiotemporal but also the frequency information and follows an end-to-end learning approach, thus providing more high-frequency information to the reconstructed video. As assessed, the proposed OVQE\_VVC significantly outperforms relevant QE methods, notably the MFQE 2.0 and STDF and achieved important 19.639\% bitrate saving when compared to the standard VVC solution. This important achievement encourages to further explore the OVQE in the future, notably by finding a better OVQE model for multi compression rates. 